\documentclass{article}
\usepackage{spconf,amsmath,amssymb,graphicx}

\def \prox{\mbox{prox}}

\def\bx{{\mathbf x}}
\def\hbx{\hat{{\mathbf x}}}
\def\by{{\mathbf y}}

\def\bn{{\mathbf n}}

\def\bz{{\mathbf z}}
\def\ba{{\mathbf a}}
\def\bs{{\mathbf s}}
\newcommand{\rR}[0]{\mathbb{R}}

\newcommand{\cL}[0]{\mathcal{L}}

\newcommand{\cS}[0]{\mathcal{S}}

\newtheorem{proposition}{Proposition}

\title{Iterative log thresholding}
\name{Dmitry Malioutov, Aleksandr Aravkin}
\address{T.J. Watson IBM Research center}
\begin{document}
%
\maketitle
\begin{abstract}
Sparse reconstruction approaches using the re-weighted $\ell_1$-penalty
have been shown, both empirically and theoretically, to provide a significant 
improvement in recovering sparse signals in comparison to the  
$\ell_1$-relaxation. However, numerical optimization of such penalties involves 
solving problems with $\ell_1$-norms in the objective many times. 
Using the direct link of reweighted $\ell_1$-penalties to the concave 
log-regularizer for sparsity, we derive a simple prox-like algorithm for 
the log-regularized formulation.
The proximal splitting step of the algorithm has a closed form solution, and we call
the algorithm {\em log-thresholding} in analogy to soft thresholding for the  
$\ell_1$-penalty. We establish convergence results, and demonstrate 
that log-thresholding provides more accurate sparse reconstructions compared to 
both soft and hard thresholding. Furthermore, the approach can be 
directly extended to optimization over matrices with penalty for rank (i.e. 
the nuclear norm penalty and its re-weigthed version), where we suggest a 
{\em singular-value log-thresholding} approach.
\end{abstract}
\section{Introduction}
\label{sec:intro}

We consider sparse reconstruction problems which attempt to find sparse
solutions to over-determined systems of equations. A basic example
of such a problem is to recover a sparse vector $\bx \in \rR^N$ from 
measurements $\by = A \bx + \bn$, where $\by \in \rR^M$ with $M < N$,
and $\bn$ captures corruption by noise. Attempting to find maximally 
sparse solutions is known to be NP-hard, so convex relaxations involving 
$\ell_1$-norms have gained unprecedented popularity. Basis pursuit 
(or LASSO in statistics literature) minimizes the following objective:
\begin{eqnarray}
\label{eqn:LASSO}
\min \Vert \by - A \bx \Vert_2^2 + \lambda \Vert \bx \Vert_1
\end{eqnarray}
Here $\lambda$ is a parameter that balances sparsity versus the norm 
of the residual error. There is truly a myriad of algorithms for solving 
(\ref{eqn:LASSO}) (see e.g.~\cite{Figueiredo2007,EwoutvandenBerg,Osher2010}), 
and for large-scale instances, variations of 
iterative soft thresholding have become very popular:
\begin{equation}
\label{eqn:ista}
\bx^{(n+1)} = \cS_{\lambda} \left( \bx^{(n)} + A^T(\by - \bx^{(n)}) \right)
\end{equation}
where $\cS_{\lambda}(\bz)$ applies soft-thresholding for each entry:
\begin{equation}
	\label{eqn:soft_thresh}
	\cS_{\lambda}(z_i) = \mbox{sign}(z_i) \max(0, |z_i| - \lambda).
\end{equation}
Based on operator splitting and proximal projection theories, the algorithm 
in (\ref{eqn:ista}) converges if the spectral norm $\Vert A \Vert < 1$
\cite{daubechies_ISTA, combettes}. This can be achieved simply by rescaling $A$. 
Accelerated versions of iterative thresholding have appeared \cite{fista}.

An exciting albeit simple improvement over $\ell_1$-norms for approximating sparsity 
involves weighting the $\ell_1$-norm: $\sum_i w_i |x_i|$ with $w_i > 0$. Ideal weights
require knowledge of the sparse solution, but a practical idea is to use weights 
based on solutions of previous iterations \cite{fazel_log_det, candes_rev_l1}:
\begin{equation}
w_i^{(n+1)} = \frac{1}{\delta + |\hat{x}^{(n)}_i|}
\end{equation}
This approach can be motivated as a local linearization of the $\log$-heuristic 
for sparsity \cite{fazel_log_det}. There is strong empirical \cite{candes_rev_l1} 
and recent theoreical evidence that reweighted $\ell_1$ approaches improve recovery 
of sparse signals, in the sense of enabling recovery from fewer measurements 
\cite{needell2009noisy, amin2009weighted}. 

In this paper, we consider the log-regularized formulation that gives 
rise to the re-weighting schemes mentioned above, and 
propose a simple prox-like optimization algorithm for its optimization. 
We derive a closed-form solution for the proximal step, which we call $\log$-thresholding.  
We establish monotone convergence of iterative $log$-thresholding (ILT) to its fixed points, 
and derive conditions under which these fixed points are 
local minima of the log-regularized objective. Sparse recovery 
performance of the method on numerical examples surpasses both soft and hard iterative thresholding (IST and IHT). 
We also extend the approach to minimizing rank for matrix functions via singular value 
$\log$-thresholding. 

To put this into context of related work, 
\cite{chartrand_lp_thresh} has considered iterative
thresholding based on non-convex $\ell_p$-norm penalties for sparsity. 
However,  these penalties do not have a connection to re-weighted $\ell_1$ optimization. 
Also, \cite{mazumder2011sparsenet} have investigated coordinate descent based 
solutions for non-convex penalties including the log-penalized penalty, 
but their approach does not use closed form log-thresholding.
Finally, general classes of non-convex penalties, their benefits for sparse recovery,
and reweighed convex-type methods for their optimization are studied in~\cite{Lin2010}.
This class of methods is different from the log-thresholding we propose.

\section{ISTA as proximal splitting}
\label{S:ista_prox_splitting}

We briefly review how soft-thresholding can be used to solve the sparse
reconstruction problem in (\ref{eqn:LASSO}). Functions of the form 
$f(x) = h(x) + g(x)$ where $h(x)$ is convex differentiable 
with a Lipschitz gradient, and $g(x)$ is general convex can be solved by a 
general proximal splitting method \cite{combettes}:
\begin{equation}
	\label{eqn:prox_split}
	\hat{\bx}^{(n+1)} = \prox_g \left( \bx^{(n)} - \nabla h(\bx^{(n)}) \right).
\end{equation}
The prox-operation is a generalization of projection onto a set to general convex 
functions:
\begin{equation}
\prox_h(\bx) = \arg \min_{\bz} h(\bz) + \frac{1}{2} \Vert \bx - \bz \Vert_2^2.
\end{equation}
If $h(\bx)$ is an indicator function for a convex set, then the prox-operation is 
equivalent to the projection onto the set, and ISTA itself is equivalent to the 
projected gradient approach. 

Forward-backward splitting can be applied to the sparse recovery problem 
(\ref{eqn:LASSO}) by deriving the proximal operator for $\ell_1$-norm, which is precisely 
the soft-thresholding operator in (\ref{eqn:soft_thresh}). The convergence of ISTA in (\ref{eqn:ista}) thus 
follows directly from the theory derived for forward-backward splitting \cite{combettes}.

\section{Log-thresholding}
\label{S:log_thresh}

The reweighted-$\ell_1$ approach can be justified as an iterative 
upper bounding by a linear approximation to the concave $\log$-heuristic 
for sparsity (here $\delta$ is a small positive constant) \cite{fazel_log_det}:
\begin{equation}
\label{eqn:log_reg}
	\min f(\bx) = \min \Vert \by - A \bx \Vert_2^2 + \lambda \sum_i \log(\delta + | x_i|).
\end{equation}
While the $\log$-penalty is concave rather than convex, we still consider 
the scalar proximal objective around a fixed $x$:
\begin{equation}
\label{eqn:log_prox}
g_\lambda (z) \triangleq (z - x)^2 + \lambda \log(\delta + |z|).
\end{equation}
We note that for $\delta$ small enough, the global minimum of $g_\lambda (z)$ 
over $z$ (with $x$ held constant) is always at $0$. However, when 
$|x| > x_0 \triangleq \sqrt{2 \lambda} - \delta$, the function also exhibits 
a local minimum, which disappears for small $x$. 
We show that  it is the local, rather than the global minimum, that provides the link 
to re-weighted $\ell_1$ minimization and is key to the log-prox algorithm we propose. 
Indeed, it is also the local (and not global) minimum that provides the link to iteratively 
re-weighteld $\ell_1$ algorithms. 

\begin{figure}[t]
  \centerline{\includegraphics[width=9.5cm]{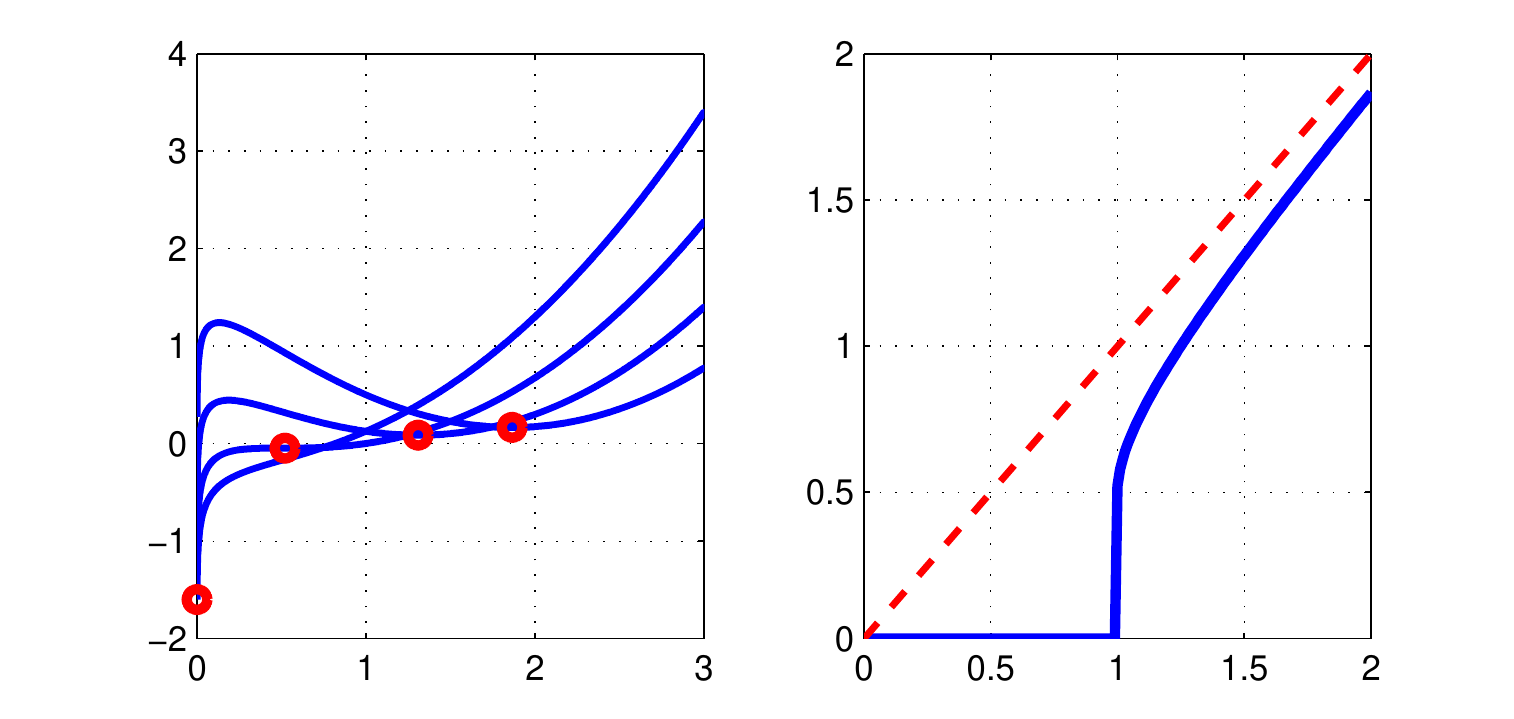}}
\caption{Illustration of log-thresholding.}
\label{fig:log_local_min}
\end{figure}

Our algorithm arises directly from first order necessary 
conditions for optimality. For $|x| > x_0$, we solve the equation $\nabla g_{\lambda} = 0$ to find the 
local minimum in closed-form. We call this operation {\em log-thresholding}
, $\cL_{\lambda}(x)$:
\begin{equation}
\label{eqn:log_thresh}
	\cL_{\lambda}(x) = \begin{cases} \frac{1}{2}\left((x_i-\delta) + \sqrt{ (x_i+\delta)^2 - 2\lambda }\right), 
			~ x > x_0 \\
		\frac{1}{2} \left( (x_i+\delta) - \sqrt{ (x_i-\delta)^2 - 2\lambda} \right), 
		~ x < - x_0\\
		0, \mbox{ otherwise} \end{cases}
\end{equation}
where $x_0 = \sqrt{2 \lambda} - \delta$. We illustrate log-thresholding 
in Figure \ref{fig:log_local_min}. The left plot 
shows $g_{\lambda} (z)$ as a function of $z$ for several values of $x$. For large
$x$ the function has a local minimum, but for small $x$ the local minimum disappears.
For $\log$-thresholding we are specifically interested in the the local 
minimum: an iterative re-weighted $\ell_1$ approach with small enough step size starting 
at $x$, i.e. beyond the local minimum, will converge to the local minimum, avoiding the 
global one. The right plot in Figure \ref{fig:log_local_min} shows the $\log$-thresholding 
operation $\cL_{\lambda}(x)$ with $x_0 = 1$ as a function of $x$. It can 
be seen as a smooth alternative falling between hard and soft thresholding.

In analogy to ISTA, we can now formally define the iterative $\log$-thresholding 
algorithm:
\begin{equation}
	\label{eqn:ILT}
	\hat{\bx}^{n+1} = \cL_{\lambda} \left( \bx^{n} + A^T(\by - A\bx^{n}) \right)
\end{equation}
where $\cL_{\lambda}(\bz)$ applied the element-wise $\log$-thresholding operation 
we obtained in (\ref{eqn:log_thresh}). We establish its convergence next.

\subsection{Convergence of iterative log-thresholding}

The theory of forward-backward splitting does not allow an analysis of
$\log$-thresholding, because the log is non-convex, and 
log-thresholding is not a contraction (in particular, it is not firmly non-expansive). 
Therefore, for the analysis  
we use an approach based on optimization transfer using surrogate 
functions \cite{Lange_surrogate} to prove convergence of ILT to its fixed points. 
At a high-level the analysis follows the program for IHT in \cite{blumensath2008iterative}, 
but some of the steps are notably different, and in particular 
some assumptions on the operator action are necessary to establish that 
fixed points correspond to local minima of our formulation. In the appendix we establish:
\begin{proposition}
\label{thm:ilt_convergence}
Under the assumption $\|A\|_2 < 1$, the ILT algorithm in (\ref{eqn:ILT}) monotonically 
decreases the objective $f(\bx)$ in (\ref{eqn:log_reg}), and converges to fixed points. 
A sufficient condition for these fixed points to be local minima is that $A$ restricted 
to the non-zero coefficients is well-conditioned, specifically that the lowest singular 
values of the restriction are greater than $\frac{1}{2}$.
\end{proposition}

\section{ Singular Value Log-thresholding}

A closely related problem to finding sparse solutions to systems of linear
equations is finding low-rank matrices from sparse observations, known as 
matrix completion: 
\begin{equation}
	\label{eqn:matrix_completion}
	\min \mbox{rank}(X) \mbox{ such that} ~
	X_{i,j} = Y_{i,j}, \{ (i,j) \in \Omega \}
\end{equation}
Similar to sparsity, rank is a combinatorial objective 
which is typically intractable to optimize directly. However, the nuclear 
norm $\Vert X \Vert_* \triangleq \sum_i \sigma_i(X)$, where $\sigma_i(X)$ are the singular values of X, 
serves as the tightest convex relaxation of rank, analogous to 
$\ell_1$-norm being the convex relaxation of the $\ell_0$-norm. In fact, 
the nuclear norm is exactly the $\ell_1$-norm of the singular value spectrum
of a matrix. This connection enables the application of various singular
value thresholding algorithms: for instance, the SVT algorithm of 
\cite{candes_SVT} alternates soft-thresholding of the singular value spectrum
with gradient descent steps. In the experimental section we investigate
a simplified singular-value log-thresholding algorithm for matrix completion, 
where we replace soft thresholding with hard and log-thresholdings. 
We present very promising empirical results of singular value log-thresholding in 
Section \ref{S:experiments}, and a full convergence analysis will appear in a later publication. 
\begin{figure}[t]
  \centerline{\includegraphics[width=8.5cm]{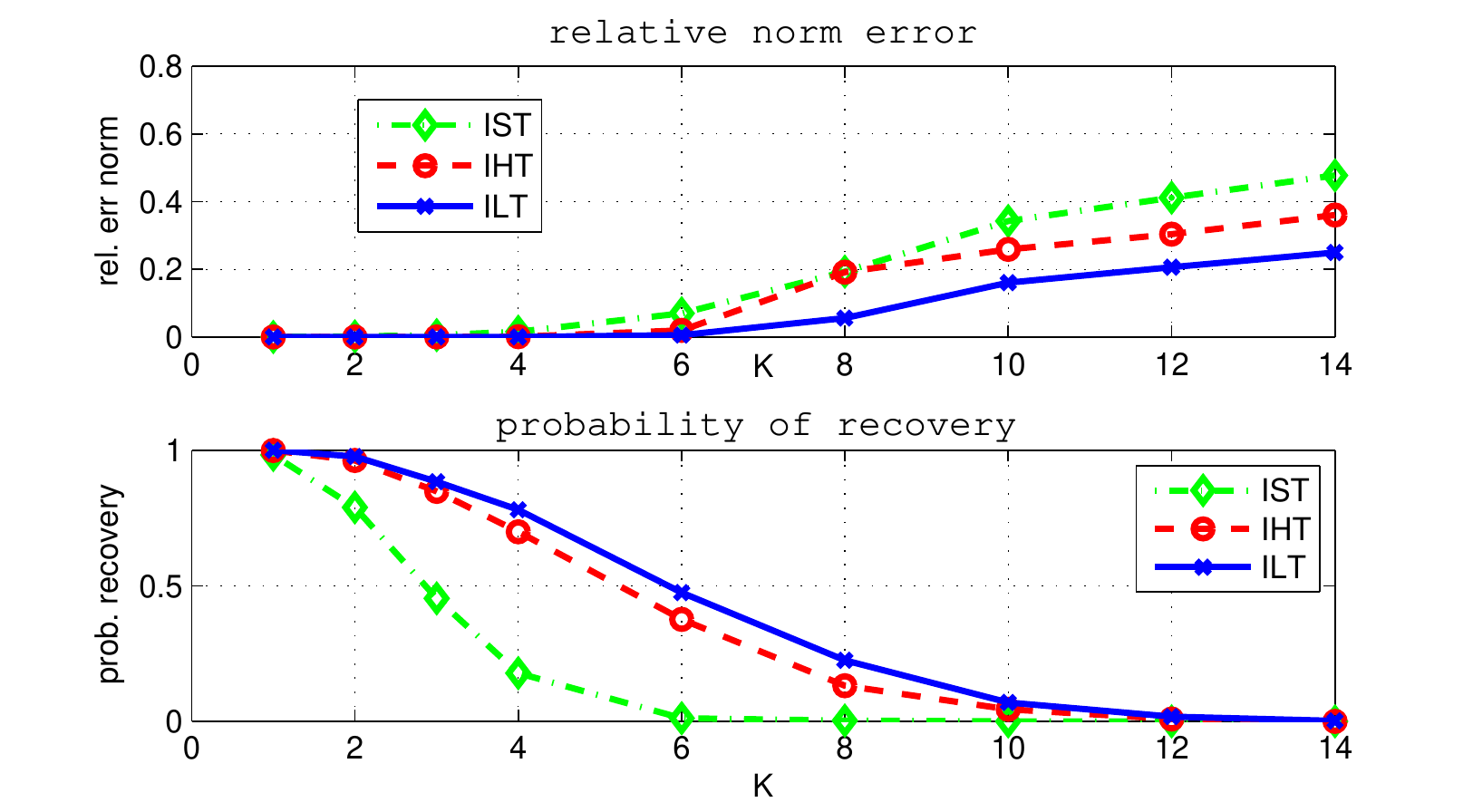}}
\caption{Noiseless sparse recovery: (a) average error-norm (b) probability of 
exact recovery after $250$ iterations over $1000$ random trials. $M = 100, N = 200$.}
\label{fig:results_no_noise}
\vspace{-0.1in}
\end{figure}

\section{Experiments}
\label{S:experiments}

We investigate the performance of iterative log thresholding via
numerical experiments on noiseless and noisy sparse recovery. Intuitively 
we expect ILT to recover sparser solution than soft-thresholding due 
to the connection to re-weighted-$\ell_1$ norms, and also to behave 
better than the non-smooth iterative hard thresholding.

First we consider sparse recovery without noise, i.e. we would like to find 
the sparsest solution that satisfies $\by = A \bx$ exactly. One could in 
principle solve a sequence of problems (\ref{eqn:LASSO}) with decreasing
$\lambda$, i.e. increasing penalty on $\Vert \by - A \bx \Vert_2^2$ via IST, 
IHT, ILT. Howeveer, when we know an upper bound $K$ on the desired number of non-zero 
coefficients, a more successful approach is to adaptively change $\lambda$ to 
eliminate all except the top-$K$ coefficients in each iteration\footnote{ This 
is easy for IST and IHT by sorting $|\bx|$ in descending order: let 
$\bs = \mbox{sort} |\bx|$ then $\lambda = s_{K+1}$. For ILT we have 
$\lambda = \frac{(x_{K+1} + \delta)^2}{4}$ from (\ref{eqn:log_thresh}).} 
as used e.g. in \cite{maleki2009coherence}. We compare the performance of IST, IHT, 
and the proposed ILT in Figure \ref{fig:results_no_noise}. We have 
$N = 200, M = 100$ and we vary $K$. Apart 
from changing the thresholding operator, all the algorithms are exactly the same.
The top plot shows the average reconstruction error from the true sparse 
solution $\Vert \hbx - \bx^* \Vert_2$. It is averaged over $1000$ trials allowing 
IST, IHT and ILT to run for up-to $250$ iterations. The bottom plot shows probability 
of recovering the true sparse solution. We can see that ILT is superior in both 
probability of recovery (higher probability of recovery) and in reconstruction error 
(lower reconstruction error) over both IST and IHT.

\begin{figure}[t]
  \centerline{\includegraphics[width=9.5cm]{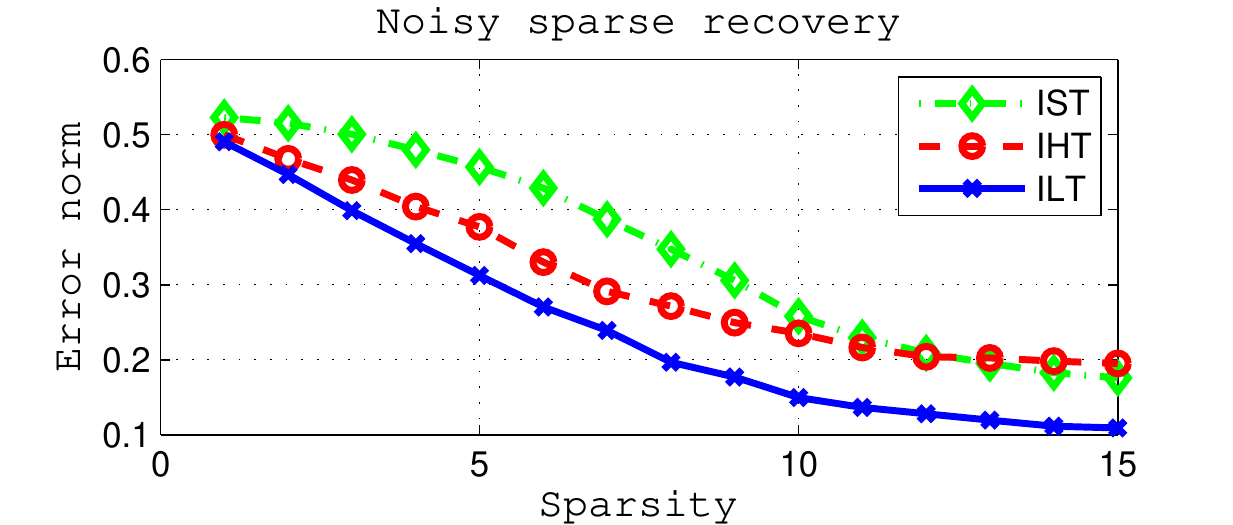}}
\caption{Sparse recovery with noise. Average error vs. sparsity over $100$ trials, 
 after $250$ iterations.}
\label{fig:results_noisy}
\vspace{-0.1in}
\end{figure}

Our next experiment compares the three iterative thresholding algorithms on noisy data. 
Since regularization parameters have a different meaning for the different penalties, 
we plot the whole solution path of squared residual error vs. sparsity for the
three algorithms in Figure \ref{fig:results_noisy}. We compute
the average residual norm for a given level of sparsity for all three algorithms, 
averaged over $100$ runs. We have $M = 100, N = 200$, $K = 10$ and a small amount 
of noise is added. We can see that the iterative log thresholding consistently 
achieves the smallest error for each level of sparsity.

\begin{figure}[t]
 \centerline{\includegraphics[width=9.5cm]{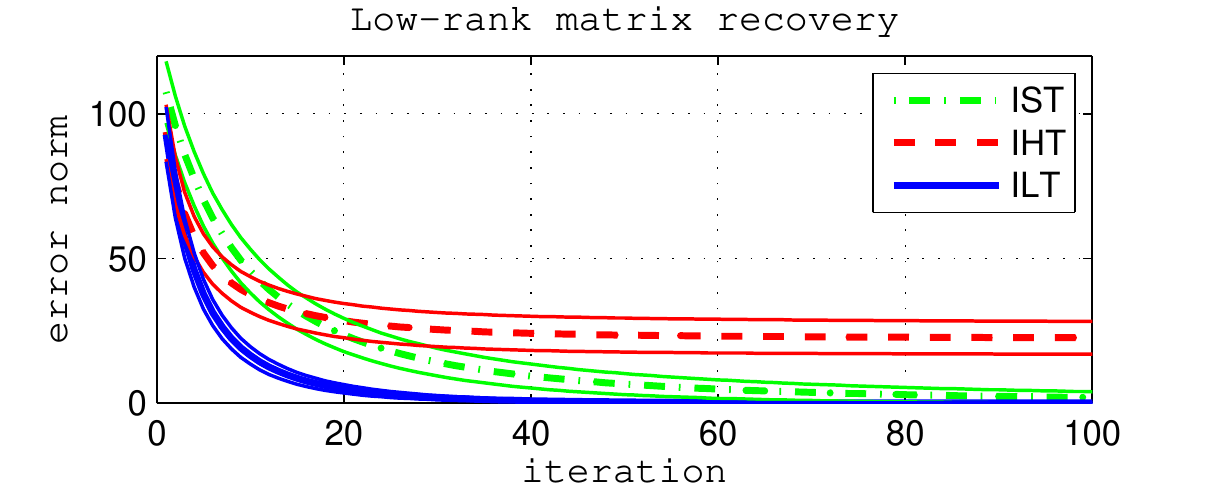}}
\caption{Illustration of singular-value log-thresholding.}
\label{fig:results_svlt}
\vspace{-0.1in}
\end{figure}

In our final experiment we consider singular value log-thresholding for matrix 
completion. We study a simplified algorithm that parallels the noiseless
sparse recovery algorithm with known number of nonzero-elements $K$. We alternate
gradient steps with steps of eliminating all but the first $K$ singular values
by soft, hard and log-thresholding. We have an $N \times N$ matrix with $30\%$ 
observed entries, $N = 100$ and rank, $K = 2$. We show the average error in Frobenius 
norm from the true underlying solution as a function of iteration number over $100$ 
random runs in Figure \ref{fig:results_svlt}. We see that the convergence of log-SV-thresholding 
to the correct solution is consistently faster.  We expect similar improvements 
to hold for other algorithms involving soft-thresholding, and to other problems beyond 
matrix completion, e.g. robust PCA.

\section{Convergence of ILT}
\label{S:ILT_convergece}

Here we establish Proposition \ref{thm:ilt_convergence}. We first define a 
surrogate function for $f(\bx)$ in (\ref{eqn:log_reg}):
\begin{eqnarray}
\label{eqn:surrogate}
\nonumber Q(\bx, \bz) = \Vert \by - A \bx \Vert_2^2 + \lambda \sum_i \log(\delta + | x_i|)+ \\
\Vert \bx - \bz \Vert_2^2 - \Vert A (\bx - \bz) \Vert_2^2
\end{eqnarray}
Note that $Q(\bx, \bx) = f(\bx)$. Simplifying (\ref{eqn:surrogate}) we have 
\begin{equation}
\label{funcQ}
 Q(\bx, \bz) =
 \sum_i \left( x_i - k_i(\bz))^2 + \lambda \log( x_i + |\delta|) \right)+ K(\bz), 
\end{equation}
where $k_i(\bz) =  z_i + a_i^T \by - a_i^T A \bz$ and $K(\bz)$ contains terms
independent of $\bx$. The optimization over $\bx$ is now separable, i.e. can be done 
independently for each coordinate. We can see that finding local minima over $\bx$ of 
$Q(\bx, \bz)$ corresponds to iterative log-thresholding.

Using this motivation for ILT, we can now prove convergence to fixed 
points of $f(\bx)$. First, we have: 
\begin{proposition}
$f(\hbx^n) = Q(\hbx^{n}, \hbx^n)$ and $Q(\hbx^{n+1}, \hbx^{n})$, 
are monotonically decreasing with iterations $n$ as long as the spectral 
norm $\Vert A \Vert_2 < 1$.
\end{proposition}
The proof parallels the IHT proof of \cite{blumensath2008iterative} using
the fact that $Q(\bx^{n+1}, \bx^n) = f(\bx^n) + \|\bx^{n+1} - \bx^n\|^2 - \|A(\bx^{n+1} - \bx^n)\|^2$,
which is independent of the thresholding used. The main difference for ILT is that 
$\hbx^{n+1}$ is not the global minimum of $Q(\bx, \hbx^{n})$ but it still holds 
that $Q(\hbx^{n+1}, \hbx^{n}) < Q(\hbx^{n}, \hbx^{n})$. $\diamond$\\ Next, we have: 
\begin{proposition}
\label{GradAnalysis}
Any fixed point of~\eqref{eqn:ILT} satisfies the following: 
\[
\begin{cases}
\ba_i^T(\by - A\bar \bx) = \frac{\lambda}{2 (\bar x_i + \delta)}  & \text{if} \quad \bar x_i > x_0\\
\ba_i^T(\by - A\bar \bx) = \frac{\lambda}{2 (\bar x_i - \delta)}  & \text{if} \quad \bar x_i < -x_0\\
|\ba_i^T(\by - A\bar \bx)| \leq x_0 &\text{otherwise}
\end{cases}
\]
In other words, if $|\bar x_i| > x_0$, then the corresponding gradient component 
satisfies local stationarity conditions for problem~\eqref{eqn:log_reg}, 
and if $|\bar x_i|  < x_0$, the gradient is bounded. 
\end{proposition}
{\it Proof:}
Given a fixed point $\bar \bx$ of~\eqref{eqn:ILT} define 
\begin{equation}
\label{si}
s_i = \ba_i^T(\by - A\bar \bx), 
\end{equation}
Suppose first that $\bar x_i + s_i > x_0$. Explicitly writing~\eqref{eqn:ILT},  
\[
\begin{aligned}
\bar x_i -s_i + \delta  &= \sqrt{ (\bar x_i + s_i+\delta)^2 - 2\lambda},
\end{aligned}
\]
squaring both sides, and simplifying, we have 
\[
\ba_i^T(\by - A\bar \bx) =  \frac{\lambda}{2(\bar x_i   +\delta)}\;,
\]
which is precisely equivalent to local optimality of~\eqref{eqn:log_reg} 
with respect to the $i$th coordinate. Otherwise, suppose $0 \leq \bar x_i + s_i < x_0$. Then we have 
$\bar x_i = 0$, and so $s_i \leq x_0$.

\begin{proposition}
For any fixed point $\bar x$ of the ILT algorithm~\eqref{eqn:log_reg} 
and any small perturbation $\|\eta\|_{\infty}< \epsilon$,
if $\delta$ is small enough, for small $\eta$ we have 
\[
Q(\bar x + \eta, \bar x) > Q(\bar x) + \|\mathcal{P}_0\eta\|^2 + \frac{3}{4}\|\mathcal{P}_1\eta\|^2\;,
\] 
where $\mathcal{P}_0$ and $\mathcal{P}_1$ denote the projections onto the zero and 
nonzero indices of $\bar x$.
The precise condition on $\delta$ is as follows:
\begin{equation}
\label{lambdaCond}
\frac{\lambda}{\delta} + 2\delta>  2\sqrt{2\lambda},
\end{equation}
\end{proposition}
{\it Proof:} This result follows by Proposition~\ref{GradAnalysis}, 
together with the proof technique of~\cite{blumensath2008iterative}[Lemma 3]. 
In particular, for any perturbation $\eta$, we can write  
$Q(\bar x + \eta, \bar x) - Q(\bar x, \bar x)$ as
\[
\begin{aligned}
& = \sum_i \left(-2\eta_is_i + \eta_i^2 + \lambda \log\left(\frac{|\bar x_i + \eta_i| + \delta}{|\bar x_i| + \delta}\right)\right)\\
\end{aligned}
\]
The above inequality is easily verified using~\eqref{funcQ}. Defining now $\Gamma_0 = \{i: \bar x_i = 0\}$ and $\Gamma_1 = \{i: \bar x_i \neq 0\}$, we can 
use the optimality properties of Proposition~\ref{GradAnalysis} to rewrite $Q(\bar x + \eta, \bar x) - Q(\bar x, \bar x)$:

\[
\begin{aligned}
\|\eta\|^2 + \sum_{i\in \Gamma_0} \left( -2\eta_is_i +  \lambda \log\left(\frac{|\eta_i| + \delta}{|\delta|}\right) \right) +\\
\sum_{i\in \Gamma_1} \left( -\frac{\eta_i\lambda}{\bar x_i + |\delta|}+ \lambda \log\left(\frac{|\bar x_i + \eta_i| + \delta}{|\bar x_i| + \delta}\right) \right)
\end{aligned}
\]

We now consider lower bounds for each of these two sums taking all $x_i \geq 0$ WLOG:

\[
\begin{aligned}
\sum_{i\in \Gamma_0} &-2\eta_is_i +  \lambda \log\left(\frac{|\eta_i| + \delta}{\delta}\right) \geq\\
& =  \sum_{i\in \Gamma_0} \lambda\log\left(1 + \frac{|\eta_i|}{\delta}\right) - 2\eta_i |x_0|\\
& = \sum_{i\in \Gamma_0} \left(\lambda\frac{|\eta_i|}{\delta} - 2|\eta_i| x_0 \right)- O(\|\eta\|^2)
\end{aligned}
\]
Given that \eqref{lambdaCond} holds, the quantity on the last line is positive for $\eta$ small enough. 
For $\Gamma_1$, we have 
\[
\begin{aligned}
&\sum_{i\in \Gamma_1} -\frac{\eta_i\lambda}{\bar x_i + \delta}+ \lambda \log\left(\frac{|\bar x_i + \eta_i| + \delta}{|\bar x_i| + \delta}\right) \geq \\
&\sum_{i\in \Gamma_1} \frac{-\eta_i\lambda+\eta_i\lambda}{\bar x_i + \delta}-\frac{1}{4} \frac{2\lambda\eta_i^2}{(|\bar x_i|+\delta)^2} \geq -\frac{1}{4} \|\eta\|_2^2
\end{aligned}
\]
where the last inequality comes from the fact that for any $i \in \Gamma_1$, we have $\frac{(\bar x_i + \delta)^2}{2\lambda} \geq 1$.


\begin{proposition}
ILT converges to its fixed points if $\Vert A \Vert_2 < 1$. 
Moreover, if the singular values of the restriction of $A$ to the $\Gamma_1$ columns
are greater than $\frac{1}{2}$, these fixed points must be the local minima of~\eqref{eqn:log_reg}.
\end{proposition}

The result follows from the proof technique of \cite{blumensath2008iterative}[Theorem 3];
in particular the sums $\sum_{n=1}^{N-1}\|x^{n+1} - x^n\|^2$  are monotonically increasing and bounded,
so the iterates $\{x_i\}$ must converge. Finally, we have 
\[
\begin{aligned}
Q(\bar \bx + \eta) &= Q(\bar \bx + \eta, \bar \bx) -\|\eta\|^2 + \|A\eta\|^2\\
&\geq Q(\bar \bx) - \frac{1}{4}\|\mathcal{P}_1\eta\|^2 + \|A\mathcal{P}_1\eta\|^2.
\end{aligned}
\]
This quantity is non-negative provided that the singular values 
of the restriction of $A$ to $\Gamma_1$ are greater than $\frac{1}{2}$.

$\square$

\bibliographystyle{IEEEbib}
\bibliography{strings_ilt,refs_ilt}

\end{document}